\documentclass[journal]{IEEEtran}

\usepackage{cite}
\usepackage{amsmath}
\usepackage{array}
\usepackage[dvipsnames]{xcolor}
\usepackage{graphicx}
\usepackage{booktabs}
\usepackage{hyperref}%

\begin{document}

\title{Can We Redesign a Shoulder Exosuit to Enhance Comfort and Usability Without Losing Assistance?}

\author{Roberto~Ferroni$^{*,1}$,
        Daniele~Filippo~Mauceri$^{*,1,2}$,
        Jacopo~Carpaneto$^{1}$,
        Alessandra~Pedrocchi$^{2}$,
        Tommaso~Proietti$^{1,3}$
\thanks{$^{*}$ These authors contributed equally.\\Corresponding author:\textit{ tommaso.proietti@santannapisa.it}\\$^1$ The Biorobotics Institute and Department of Excellence in Robotics \& AI, Scuola Superiore Sant’Anna, Pisa, Italy.\\$^2$ Dept. of Biomedical Engineering, Politecnico di Milano, Milano, Italy.\\$^3$ Modular Implantable Neuroprostheses (MINE) Laboratory, Università Vita-Salute San Raffaele, Milano, Italy.}
}

\maketitle

\begin{abstract}
Reduced shoulder mobility limits upper-limb function and the performance of activities of daily living across a wide range of conditions. Wearable exosuits have shown promise in assisting arm elevation, reducing muscle effort, and supporting functional movements; however, comfort is rarely prioritized as an explicit design objective, despite it strongly affects real-life, long-term usage. This study presents a redesigned soft shoulder exosuit (Soft Shoulder v2) developed to address comfort-related limitations identified in our previous version, while preserving assistive performance. In parallel, assistance was also improved, shifting from the coronal plane (shoulder abduction) to the sagittal plane (shoulder flexion) to better support functionally relevant hand positioning. A controlled comparison between the previous (v1) and redesigned (v2) modules was conducted in eight healthy participants, who performed static holding, dynamic lifting, and a functional pick and place task. Muscle activity, kinematics, and user-reported outcomes were assessed. Both versions increased endurance time, reduced deltoid activation, and preserved transparency during unpowered shoulder elevation. However, the difference between them emerged most clearly during functional tasks and comfort evaluation. The redesigned module facilitated forward arm positioning and increased transverse plane mobility by up to 30°, without increasing muscular demand. User-reported outcomes further indicated a substantial improvement in wearability, with markedly lower perceived pressure and higher ratings in effectiveness (+23\%), ease of use (+20\%), and comfort (+17\%) compared to the previous design. Taken together, these findings show that targeted, user-centered design refinements can improve comfort and functional interaction without compromising assistive performance, advancing the development of soft exosuits suitable for prolonged and daily use.
\end{abstract}


\IEEEpeerreviewmaketitle

\section{Introduction}
\IEEEPARstart{T}{he} shoulder plays a central role in upper-limb manipulation and activities of daily living (ADLs). Decreases in shoulder range of motion (ROM) impedes the capacity to perform ADLs, which can occur at all ages, as a consequence of a broad variety of conditions or injuries, such as stroke, spinal cord injury (SCI), or amyotrophic lateral sclerosis (ALS) \cite{1}. These conditions markedly affect autonomy and social participation, while increasing healthcare costs and caregiver burden \cite{2,3}.

Motivated by this, in the last three decades a large body of research in rehabilitation and assistive robotics aimed at mitigating movement deficits. The common efforts resulted in the creation of multiple upper-limb exoskeletons, whether rigid or driven by soft actuation strategies, achieving promising results in a variety of application scenarios \cite{4,5}. However, while the field has evolved, many “classical” challenges still need to be tackled, e.g. rigid architectures facing overweight and bulkiness, while soft ones being limited by complex control and sensing strategy.  One common issue to any wearable robotic architecture, perhaps the most overlooked, is comfort. Despite comfort is often being declared as a design goal or assumed to be an inherent property, especially in soft exosuits \cite{6,7}, this feature is rarely evaluated as an outcome and central to the design process.  

For rigid robots, several studies proposed mechanisms to adapt the exoskeleton’s kinematics to the user \cite{8,9}, but this remains an open challenge due to limitations in structural adjustability \cite{10}. Complementary efforts also attempted to quantify physical interaction at the interface, with few studies adopting pressure sensing approaches \cite{11}. For soft exosuits, metrics related to user experience are reported less frequently than other outcomes (e.g., reduction of muscle activity), with perceived discomfort most commonly assessed via questionnaires \cite{12,13,14}.

Yet, comfort-related issues are repeatedly acknowledged as critical for wearability and long-term use. For instance, commonly reported limitations include textile slippage, pressure-related discomfort, actuation perturbations due to muscle volume variability, and joint misalignment during motion \cite{15}. These effects may be acceptable in short-term applications, but require detailed analysis for longer-term usage \cite{16,17}. Importantly, improving wearability can generally lead to reduced generated torque \cite{18}, underscoring the challenge of achieving both assistance performance and comfort.

\begin{figure*}[!t]
    \centering
    \includegraphics[width=\textwidth]{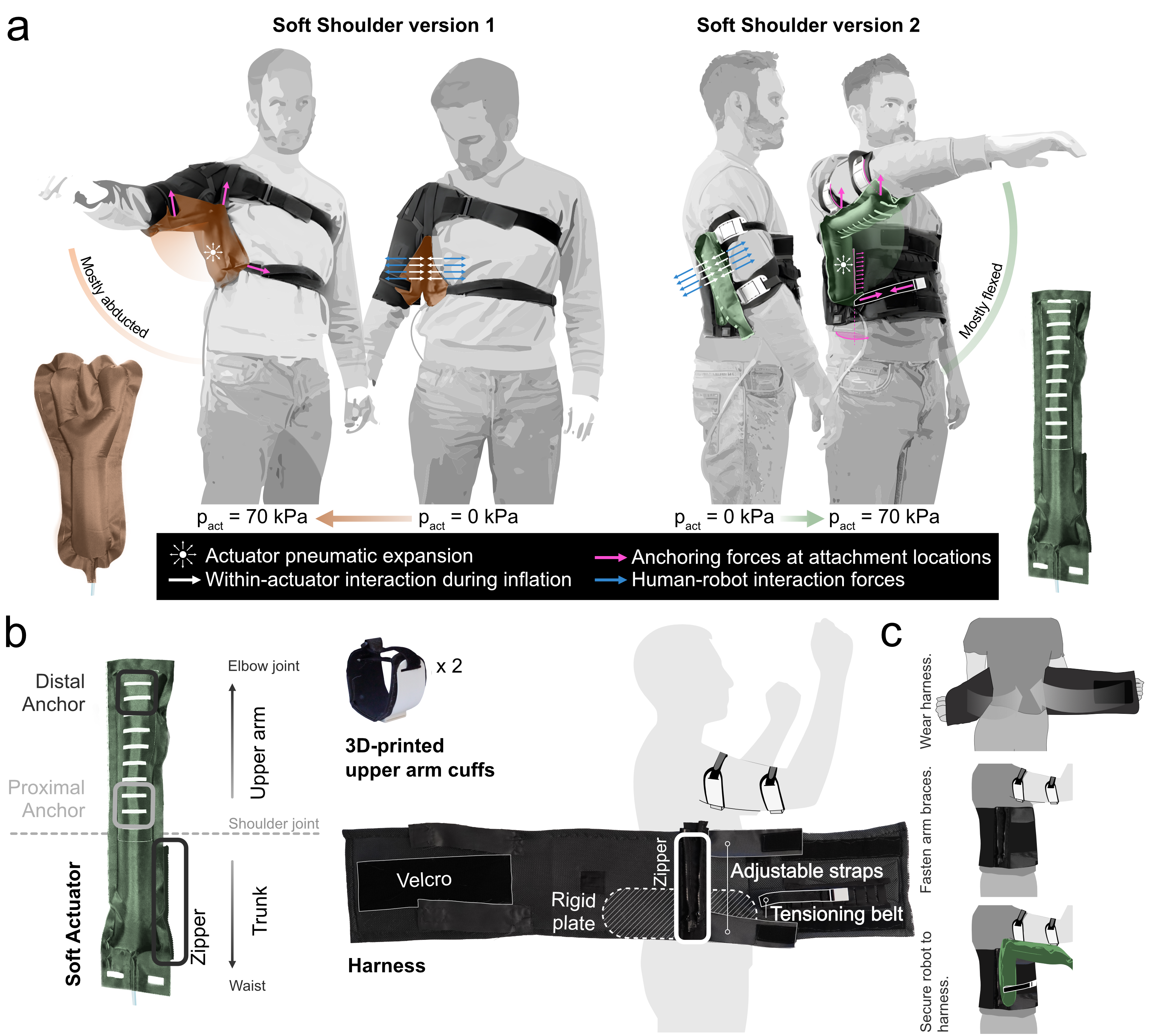}
    \caption{\textbf{System overview.} a) Comparative view of the Soft Shoulder version v1 (orange hue) and the Soft Shoulder version v2 (green hue). Each actuator consists of a fabric-based inflatable air chamber and is shown both alone and worn on a user, either powered off (pact = 0 kPa) or active to its upper pressure limit (pact = 70 kPa). Force diagrams provide a qualitative visualization of the working principle: as the actuator pressure increases, the two portions spanning the armpit (anchored respectively to the arm and the torso) begin to press against each other (white arrows), generating the initial driving force for the movement (blue arrows). As inflation progresses, the actuator chamber stiffens and approaches its straight configuration, resulting in an upward motion of the arm through the attachments that secure the actuator’s mobile portion to the upper arm. The anchoring forces at the attachment locations (purple arrows) are crucial for establishing a stable interaction with the body and ensuring that the intended primary direction of motion is consistently elicited (i.e., shoulder abduction for version 1 and shoulder flexion for version 2). b) Isolated view of each component of the renovated Soft Shoulder version 2: the soft actuator, the textile harness, and the 3D-printed upper arm cuffs. C)  Step by step donning of the device.}
    \label{fig_overview}
\end{figure*}

Beyond qualitative tools, more quantitative and interface-focused works exist. These included systematically measuring regional force and pressure limits at the body–device interface to map comfort boundaries \cite{19} and their dependence on use and loading dynamics \cite{20}; optimizing how assistive forces are routed and distributed across multiple anchoring points using human–robot models and structural design methods to reduce local interaction forces \cite{21}; and adopting broader evaluation frameworks that couple power transmission metrics with additional interface phenomena such as compression, shear, slippage, and perceived comfort to better capture overall human–exosuit interaction quality \cite{22,23}.

Building on this context, the present study moves from our previous design (“Soft Shoulder v1”), whose efficacy has been extensively demonstrated through studies on both healthy and impaired participants \cite{24,25,26}, toward a new design to address identified limitations – including comfort – by making them explicit requirement of the “Soft Shoulder v2” module. 
Besides comfort, the newer version of the robot prioritized shoulder assistance in the sagittal plane (shoulder flexion) rather than coronal plane (shoulder abduction), with the prospective goal of better positioning the hand within a workspace relevant for ADLs (e.g., personal care, feeding). 

Regarding comfort, the study combined (1) objective measures, including mechanical transparency not only during assisted shoulder elevation but also in the transverse plane to assess possible movement restrictions in shoulder horizontal adduction, with (2) direct user-reported feedback, including reported satisfaction, perceived pressure intensity on the body, and the perceived direction of shoulder motion when assistance was active.

To enable a controlled comparison between the previous (v1) and redesigned (v2) exosuit versions, we evaluated both systems in a cohort of N = 8 healthy participants, naïve to either version of the device, across multiple experimental tasks, including static holding, dynamic lifting, and a functional pick and place activity.

\section{Materials and Methods}

\subsection{Soft Shoulder version 1}
The first version of the shoulder assistive exosuit has been previously described in detail, both as a standalone device and in combination with other upper-limb assistance modules \cite{24,25,26,27}. It features a fabric-based pneumatic actuator coupled to the body through a custom shoulder brace and Velcro straps to ensure stable fitting on the torso and arm, see Fig. \ref{fig_overview}. Upon inflation, the lower portion of the actuator is constrained by the torso, while the upper portion is free to expand. This asymmetric mechanical constraint produced an upward and lateral displacement of the actuator, generating a lifting moment about the shoulder joint primarily aligned with the frontal plane, thus eliciting shoulder abduction. 

\subsection{Soft Shoulder version 2}
The redesigned shoulder exosuit was conceived to reproduce the mechanical interaction exploited in the previous version but reoriented within the sagittal plane to better support shoulder flexion, see Fig. \ref{fig_overview}. Moreover, trunk compression due to the inflation of the actuator has been often described as the main source of discomfort, especially by healthy users. To this end, both the anchoring configuration and the direction of actuator expansion were redefined to generate a forward and upward assistive force aligned with the desired movement, without relying on the torso as a lateral anatomical constraint. The new architecture introduces a set of coordinated mechanical constraints to maintain the generated torque and to stabilize and guide the actuator deformation during inflation.

Specifically, the actuator is laterally aligned along the torso via a zipper interface integrated into the harness, while 3D printed arm cuffs provide proximal and distal anchoring to the upper arm. Crucially, an inferior mechanical constraint is introduced through a belt-and-buckle system connecting the lower end of the actuator to the thoracic harness. This inferior anchoring stabilizes the actuator’s base, allowing pressurization to generate a controlled forward and upward displacement of the upper section. As a result, the actuator produces a flexion-assistive moment about the shoulder joint.

The anchoring interface consists of a belt-like waist vest designed to couple the actuator to the body while maintaining comfort and adaptability to different body sizes. The harness features a hybrid textile construction that combines a 3 mm thick 3D spacer mesh for breathability, with a high-strength non-stretch fabric providing structural support. The main structure comprises a rectangular thoracic band (22.3 × 111.5 cm) secured around the torso using two primary Velcro straps. Two additional pairs of Velcro straps tension the non-stretch textile components and allow side-by-side or overlapping configurations to accommodate different torso sizes. For improved comfort, two body-facing fabric pockets house a 1 mm thick flexible thermoplastic plate to distribute contact pressure, thereby mitigating localized pressure concentrations.

\subsection{Control unit and pressure regulation}
The pneumatic actuators were pressure-controlled using a custom wall-powered portable control unit, previously developed and validated by our group \cite{26}. Briefly, the system comprises two miniature air pumps operating in parallel, four miniature normally closed solenoid valves for inflation and deflation, and an additional normally closed solenoid valve for pump exhaust, a master microcontroller (Feather M0 WiFi, Adafruit), and custom electronics. Actuator pressure is regulated through a closed-loop bang-bang pressure control scheme with hysteresis, running at 200 Hz, based on pressure sensor readings monitoring the internal actuator pressure. Human limb kinematic data were acquired using three 9-axis inertial measurement units (IMUs, BNO055, Adafruit), managed by a second, battery-powered microcontroller (Feather M0 WiFi, Adafruit) transmitting data at 100 Hz to the master microcontroller via a TCP/IP connection. The IMUs were housed in dedicated 3D-printed PLA enclosures and placed on the user’s torso, upper arm and forearm after performing the manufacturer’s default calibration routines. The master microcontroller streamed data to a custom PyQt-based graphical user interface (GUI), which provided real-time visualization of sensor data and served as visual feedback for participants during tasks involving precise angle-tracking. 

\subsection{Study design}
The purpose of this study is to iterate the design of a soft shoulder robot, particularly in terms of user perceived comfort and functionality of the support provided. To this aim, we carried experiments with eight healthy individuals (inclusion criteria: 18-65 years old, capable of lifting 4 kg weight with their dominant hand, and with no record of major injuries at the upper limb), using both versions of the shoulder robot, in a randomized order. Each participant voluntarily joined a single testing session, where informed consent was obtained prior to participation. Data collection took place at the Biorobotics Institute of Scuola Superiore Sant’Anna, Pontedera (Italy) after approval of the testing protocol (01/2024) by the Joint Ethics Committee of the Scuola Normale Superiore and the Scuola Superiore Sant’Anna. 
To assess and compare robotic support across the two versions of the shoulder exosuit, surface electromyographic (sEMG) activity was recorded using wet electrodes (Sessantaquattro+, OT Bioelettronica) from the anterior, middle, and posterior heads of the deltoid muscle, hereafter referred to as ‘AD’, ‘MD’, and ‘PD’, respectively. Electrode placement followed the recommendations for non-invasive assessment of muscle activity using surface EMG \cite{28}. To enable inter-subject comparisons, maximum voluntary contraction (MVC) was recorded for each EMG channel at the beginning of each testing session. MVC was computed as the mean of the maximum EMG values obtained from three consecutive isometric contractions. Upper-limb kinematics were additionally measured using three IMUs placed on torso (upper third of the sternum), upper arm, and forearm, following the standard sensors calibration routine. All trials were video-recorded using a camera (Hero10, GoPro) for later verification during data processing.

\subsection{Experimental protocol}
Every participant took part in a series of activities involving the use of both versions of the shoulder exosuit (“v1”, “v2”), tested either unpowered (“robot OFF”) or providing assistance (“robot ON”). To assess mechanical transparency, participants additionally performed the same shoulder movements without wearing any robotic device, providing a free-movement baseline. The experimental protocol included two types of activities: (1) user-reported evaluations, providing qualitative feedback on comfort, perceived assistance, and satisfaction; and (2) functional assessments, providing quantitative measures of the exosuits’ effectiveness in supporting performance of physical exercises. Both exosuit versions and every tested assistance condition were randomized across participants and activities, and a rest period of 5 minutes was ensured to prevent excessive fatigue.

\subsubsection{User-reported comfort evaluation and perceived direction of assistance}
We hypothesized that the renewed version of the shoulder exosuit would provide a more comfortable wearing experience. To evaluate this, each participant was instructed to maintain a relaxed standing posture and to avoid actively contributing to the movement. A passive mobilization of the arm was then performed through a full inflation-venting cycle. Immediately after this first interaction with the robot, a diagram of the upper body was shown, and participants were asked to: (1) identify any perceived pressure spots by marking the corresponding areas on the diagram; (2) rate the intensity of each pressure sensation by selecting one of the following options: “light pressure”, “uncomfortable”, or “painful”; and (3) indicate the perceived primary direction of arm motion during assistance, choosing between “front”, “side”, or “oblique”.

\subsubsection{Comparison of static assistance}
To compare the effectiveness of exosuit assistance across the two exosuit versions during sustained physical effort, participants performed a static holding task while maintaining an external load corresponding to 2.5\% of their body mass. The task was executed in two different postures: (1) shoulder abduction, with the arm elevated to 90 degrees and directed laterally; and (2) shoulder flexion, with the arm elevated to 90 degrees and directed forward. Exosuit versions and assistance conditions were randomized across participants, and a rest period of 3 minutes was provided between repetitions to allow recovery from sustained isometric contraction. Each trial was terminated when one of the following conditions was met: (1) the participant voluntarily stopped due to fatigue; (2) the shoulder elevation angle decreased by more than 10 degrees; (3) the posture was maintained for longer than 10 minutes. Endurance time was used as the primary outcome measure, and sEMG to evaluate and compare robotic support across exosuit versions and conditions. 

\subsubsection{Comparison of dynamic assistance}
To compare movement performance across exosuit versions and assistance conditions, participants performed controlled lifting tasks while supporting an external load corresponding to 2.5\% of their body mass, following a predefined temporal trajectory displayed on a monitor. Each repetition consisted of three phases: (1) a 7 s lifting phase, during which the arm was raised from rest to 90° of elevation; (2) a 2 s hold phase at the target angle; and (3) a 7 s lowering phase returning to the initial position. Movements were performed along three directions: (1) shoulder abduction, (2) shoulder flexion, and (3) contralateral diagonal lift, moving the hand from the ipsilateral hip toward the contralateral shoulder. For each movement direction, the dynamic task was repeated three times under both unpowered (robot OFF) and assisted (robot ON) conditions for each version. Exosuit version and assistance condition were randomized across participants, and a rest period of 2 minutes was provided across repetitions along different movement directions to allow recovery from sustained dynamic weightlifting. sEMG signals were recorded as the primary outcome measure for this exercise.

\subsubsection{Mechanical transparency assessment}
The previous version of the exosuit has been shown to be mechanically transparent during shoulder elevation, meaning that it does not restrict free movement when worn unpowered \cite{24}. To confirm the mechanical transparency of the newer shoulder module and compare it with the previous version, participants were instructed to perform isolated shoulder movements along three directions: (1) abduction, raising the arm in the frontal plane from rest along the torso to the maximal achievable height; (2) flexion, raising the arm in the sagittal plane from rest along the torso to the maximal achievable height; (3) horizontal adduction, rotating the arm from the ipsilateral to the contralateral side in the horizontal plane. For shoulder abduction and flexion, movements were performed under three conditions: no robot, v1 unpowered (v1 OFF), and v2 unpowered (v2 OFF), allowing assessment of mechanical transparency during elevation. For shoulder horizontal adduction, movements were performed under no robot with external arm-weight support, v1 powered (v1 ON), and v2 powered (v2 ON), as this task required continuous arm-weight compensation. In the no robot condition, arm weight was supported either by exosuit when powered or by a low-friction surface that allowed free rotation with negligible resistance. For each shoulder movement and condition, participants completed three consecutive repetitions, without any rest periods, as the task involved no external loading and were not expected to induce significant fatigue. Shoulder ROM and sEMG data were recorded and compared across conditions.

\subsubsection{Tests of functional support during pick and place task}
To evaluate the naturalness of upper-limb movements enabled by the two shoulder exosuit versions during a functional, goal-oriented task, participants performed a repetitive pick and place activity while seated at a desk, inspired by the box-and-block-test. The task required moving as many wooden cubic blocks (2.5 cm side length) as possible from a starting area to a target area, both defined by tape-marked squares on the table surface, while clearing a rectangular obstacle (15 cm in height) positioned between them. The starting area was located on the ipsilateral side of the arm wearing the exosuit, at 45° lateral to the participant’s torso, whereas the target area was positioned toward the contralateral shoulder at 45° in the opposite direction. Each trial was performed with both exosuit versions providing assistance (v1 ON, v2 ON), with the order randomized across participants. As the task primarily involved rotation of the arm in the horizontal plane, IMU data were recorded to assess torso and arm ROMs as primary outcome variables, while sEMG was collected to quantify robotic support.

\subsubsection{Post-experiment user satisfaction assessment}
At the end of the experimental session, participants completed a questionnaire based on the Quebec User Evaluation of Satisfaction with Assistive Technology (QUEST). Of the original 12-item questionnaire, only the 8 product-related items were administered, as they directly reflect user satisfaction with the device itself. These items assessed size, weight, adjustability, safety, durability, ease of use, comfort, and effectiveness, and were rated on the standard QUEST 0–5 satisfaction scale. The four service-related items were not included, as they concern aspects such as technical support and service delivery, which were not relevant to the evaluation of robotic prototypes tested under controlled laboratory conditions.

\begin{figure*}[!t]
    \centering
    \includegraphics[width=\textwidth]{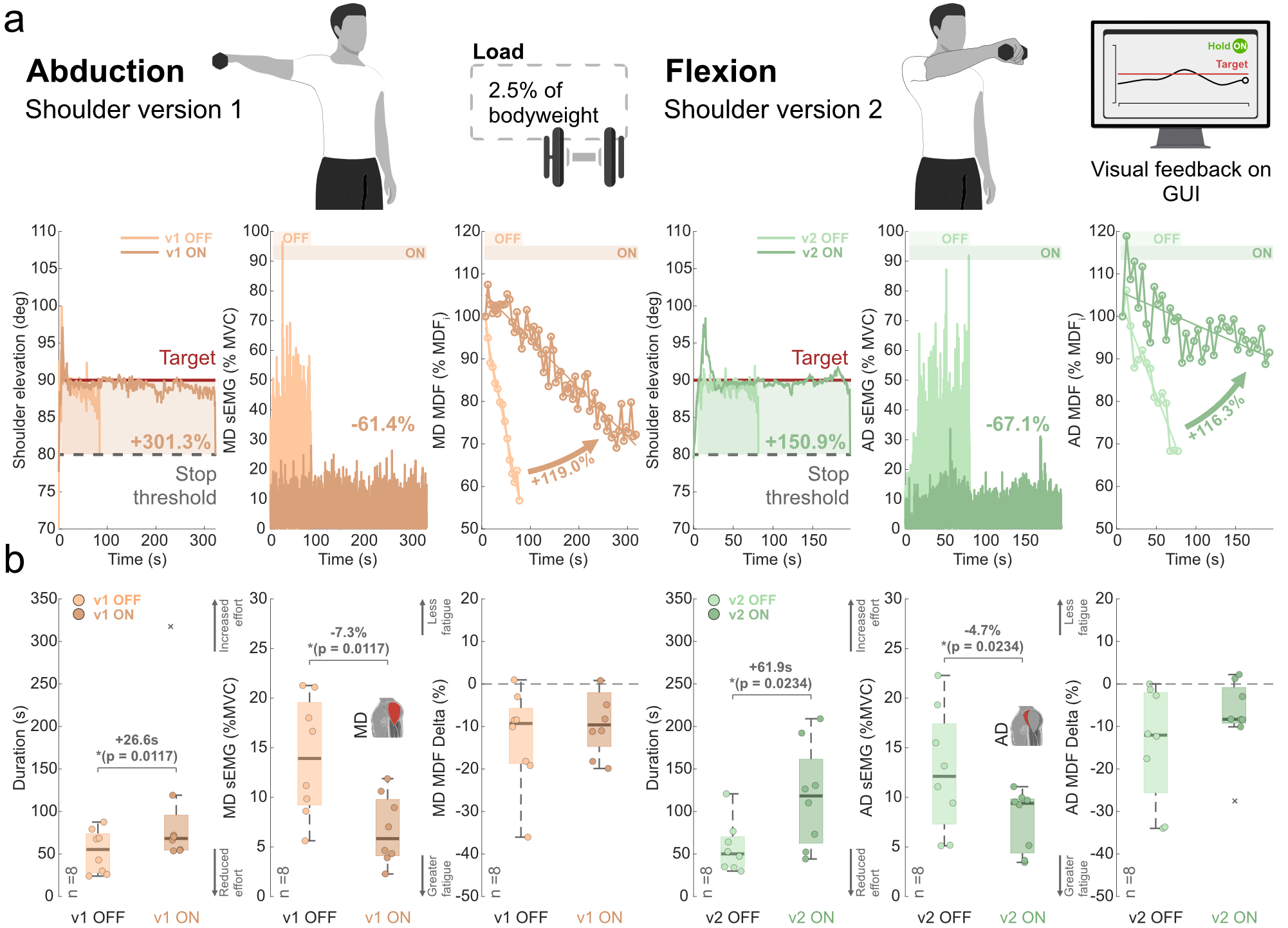}
    \caption{\textbf{Static holding performance during target movements of shoulder exosuit v1 and v2.} a, Representative time-series from a single participant (S6) during sustained static holding tasks. From left to right, shoulder elevation angle, normalized sEMG activity, and MDF are shown for the primary assisted movement of each exosuit version. The first three traces refer to shoulder abduction with exosuit version 1, and the last three traces to shoulder flexion with version 2, each shown in unassisted (robot OFF) and assisted (robot ON) conditions. Horizontal reference lines in the elevation traces indicate the target angle (90°) and the stopping threshold (80°). Across both movements, assisted conditions are associated with longer endurance, reduced muscle activation, and a slower decline in MDF. b, Group level summary of endurance time, normalized sEMG activity, and MDF$\Delta$ for each exosuit version in its target movement. Both versions improved task duration and reduced activation of the primary agonist muscle, while fatigue-related metrics exhibited consistent but non-significant trends toward slower decline under assistance.}
    \label{fig-static}
\end{figure*}

\subsubsection{Data processing} 
sEMG signals were processed offline to extract both amplitude-based metrics to assess muscle activation and, additionally, frequency-domain indices related to muscle fatigue during sustained isometric tasks. For static holding tasks, sEMG analyses were performed over a subject-specific activation window defined as the minimum task duration across all conditions and exosuit versions, to ensure equal-duration comparisons.

Raw sEMG signals, sampled at 2 kHz, were first band-pass filtered (fourth order, 10-400 Hz) and notch filtered at 50 Hz to remove power-line interference. The filtered signals were full-wave rectified and low-pass filtered (fourth order, 10 Hz) to obtain the envelope, which was then normalized relative to the MVC of each participant’s recorded muscles (which underwent the same pre-processing). Finally, the median of the normalized sEMG envelope was computed over the selected activation window, as a robust estimate of muscle activation.

Muscle fatigue was further quantified using the median frequency (MDF) of the sEMG power spectrum. Following band-pass and notch filtering, the sEMG segment corresponding to the static holding phase (activation window, as defined above) was extracted and divided into consecutive 5 s epochs. Within each epoch, a 4 s sliding window with 75\% overlap was applied. Prior to spectral analysis, a Hanning window was used, and the power spectrum was computed via fast Fourier transform. Outliers in the resulting MDF time-series were corrected using a Hampel identifier filter (using 25 surrounding samples per side) \cite{29}. The MDF time-series was normalized to the initial MDF value and expressed as a percentage, such that the first sample corresponded to 100\%. A linear trend was estimated using ordinary least squares regression, with a negative slope indicating the development of muscle fatigue over time \cite{30}. In addition, the relative change in MDF (MDF$\Delta$) between the beginning and the end of the activation was computed. Specifically, MDF$\Delta$ was calculated as the relative difference between the median MDF of the last two samples (MDF$_f$) and the median MDF of the first two samples (MDF$_i$), using MDF$_i$ as reference. Negative MDF$\Delta$ values indicate a decrease in median frequency and are associated with increased muscle fatigue \cite{29}.

Upper-limb kinematics were estimated from the absolute orientations of the IMUs placed on the torso, upper arm and forearm. IMU quaternions were converted to rotation matrices to compute relative segment orientations. Shoulder elevation (sEL) was obtained from the angle between the upper arm longitudinal axis and the vertical direction; shoulder azimuth (sAZ) from the relative rotation between torso and upper arm, on the transverse plane; elbow flexion-extension (eFE) from the angle between the longitudinal axes of the upper arm and forearm.

Concerning the user-reported comfort evaluation, individual pressure maps from each participant were spatially registered to a common torso template. These maps were segmented to obtain binary masks corresponding to each intensity level and each predefined anatomical region (upper arm, armpit, flank, torso), identified on the template. For each participant and anatomical region, a weighted pressure score was computed by combining the area of pixels assigned to each intensity level with integer intensity weights (1: light pressure, 2: uncomfortable, 3: painful). These regional scores were then averaged across participants to obtain mean intensity maps for each exosuit version. This procedure enabled direct comparison of spatial pressure distribution and perceived contact intensity across conditions on a common anatomical reference frame.

\subsection{Statistical analysis}
Statistical analyses were performed using custom scripts in MATLAB (The Mathworks, Natick, MA). Descriptive statistics are reported as medians and interquartile ranges (IQR). To assess the effect of the robotic assistance conditions, a non-parametric repeated-measures approach was adopted, given the sample size and the within-subject study design. Analyses were conducted separately for each movement direction. For comparisons across more than two experimental conditions (e.g., robot OFF, v1 OFF, v2 OFF), a Friedman test was used to evaluate omnibus effects. When a significant main effect was observed, or when comparisons involved only two paired conditions, post-hoc analyses were performed using Wilcoxon signed-rank tests. To account for multiple comparisons across outcome variables, p-values were adjusted using the Benjamini-Hochberg procedure to control the false discovery rate (FDR). Statistical significance was defined as an FDR-corrected p-value $< 0.05$. Effect sizes for pairwise comparisons were computed as $r=Z/\sqrt{N}$, where N is the number of paired observations, and interpreted as small (0.1), medium (0.3), or large ($\geq0.5$).

\section{RESULTS}

\subsection{Both shoulder exosuit versions deliver effective support in their target movements}
We hypothesized that the redesigned shoulder exosuit could achieve performance comparable to the previous module while supporting the upper arm during sustained holding tasks. Fig.2 summarizes the results of the static exercise for each exosuit version, tested in the primary movement targeted by each design (shoulder abduction for version 1, shoulder flexion for version 2). Fig. \ref{fig-static}.a reports representative time-series from an exemplary participant (S6), comparing unassisted and assisted conditions. In the assisted condition, endurance time increased, muscle activation reduced, and the temporal evolution of fatigue-related indices appeared slower. Fig. \ref{fig-static}b provides a group level overview of endurance time, normalized sEMG activity, and the MDF$\Delta$ index as a proxy of muscle fatigue. Both exosuit versions consistently enhanced participants’ performance in their respective movement planes. During shoulder abduction, version 1 increased endurance time by 26.6 s (95\% CI 4.4 s to 31.6 s, p$_{FDR}$ = 0.0117, r = 0.63, n = 8), accompanied by a significant reduction in MD activation (median $\Delta$ = -7.3\%, 95\% CI -10.5\% to -6.0\%, p$_{FDR}$ = 0.0117, r = 0.63, n = 8). Similarly, during shoulder flexion, version 2 increased endurance time by 61.9 s (95\% CI 14.2 s to 92.5 s, p$_{FDR}$ = 0.0234, r = 0.63, n = 8) and reduced AD activation (median $\Delta$: -4.7\%, 95\% CI -8.3\% to -0.2\%, p$_{FDR}$ = 0.0234, r = 0.59, n = 8). Regarding muscle fatigue, MDF$\Delta$ values showed a consistent tendency toward less negative slopes in the assisted condition, suggesting a slower shift toward lower frequencies. However, these trends did not reach statistical significance at the group level. This result likely reflects the combined effect of the limited sample size and the conservative time window selection, which enforced equal-duration comparison windows across versions and conditions.

\begin{figure*}[!t]
    \centering
    \includegraphics[width=\textwidth]{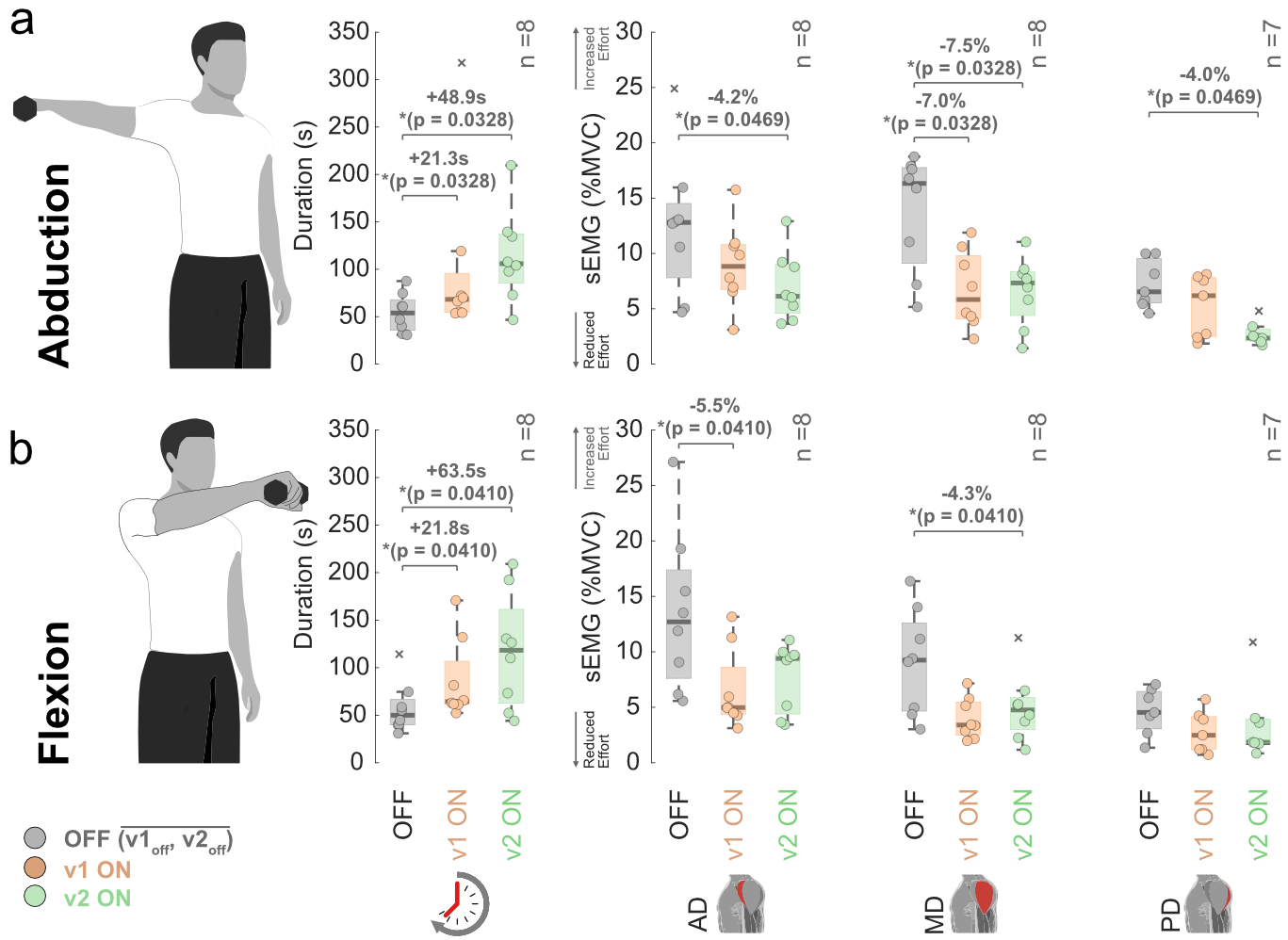}
    \caption{\textbf{Comparison of exosuit versions in static task performance.} a, Boxplots show endurance time and normalized sEMG activity (\%MVC) of the AD, MD and PD muscles for the three conditions (robot OFF, v1 ON, v2 ON). Both exosuit versions increased endurance time relative to baseline, with version 1 primarily reducing MD activation and version 2 producing broader reductions across deltoid heads. b, Endurance time and deltoid sEMG activity are shown for the same tested conditions. Both versions enhanced task duration, while exhibiting muscle-specific support patterns, with version 1 reducing AD activity and version 2 reducing MD activity.}
    \label{fig-static2}
\end{figure*}

\subsection{Comparable static task performance across exosuit versions }
To further assess the effectiveness of the two exosuit versions during each shoulder movement direction, endurance time and deltoid muscle activity were compared across three assistance conditions (robot OFF, v1 ON, v2 ON). The robot OFF condition was obtained by averaging the unassisted trials performed with exosuit version 1 and version 2, after preliminary statistical testing confirmed that no significant differences were present between v1 OFF and v2 OFF, for any outcome variable. 

During abduction (Fig. \ref{fig-static2}.a), the Friedman test revealed a significant effect of assistance condition on endurance time ($\chi^2$(2) = 13.0, p=0.0015). Post-hoc Wilcoxon tests showed that both exosuit versions significantly increased task duration compared to baseline (v1 ON: median $\Delta$ = +21.3 s, 95\% CI 11.4 s to 31.6 s, p$_{FDR}$ = 0.0328, r = 0.63, n = 8; v2 ON: median $\Delta$ = +48.9 s, 95\% CI 33.1 s to 78.6 s, p$_{FDR}$ = 0.0328, r = 0.63, n = 8), while no significant difference was observed between the two assisted conditions (p$_{FDR} >$  0.05).

A significant main effect of assistance condition was also observed for deltoid muscle activity. For the MD, the primary muscle involved in shoulder abduction, the Friedman test indicated a significant effect ($\chi^2$(2) = 12.3, p=0.0022). Both v1 ON (median $\Delta$ = -7.0\%, 95\% CI -10.9\% to -4.9\%, p$_{FDR}$ = 0.0328, r = 0.63, n = 8) and v2 ON (median $\Delta$ = -7.5\%, 95\% CI -9.4\% to -5.2\%, p$_{FDR}$ = 0.0328, r = 0.63, n = 8) significantly reduced MD activation relative to baseline, with no difference between the two assisted conditions.

In contrast, reductions in AD and PD activity were selectively observed with v2 ON. The Friedman test revealed significant effects for both AD ($\chi^2$(2) = 6.3, p=0.044) and PD ($\chi^2$(2) = 10.3, p=0.0058) activity. Post-hoc analyses revealed that v2 ON significantly reduced activation of both muscles compared to baseline (AD: median $\Delta$ = -4.2\%, 95\% CI -9.7\% to -0.8\%, p$_{FDR}$ = 0.0469, r = 0.60, n = 8; PD: median $\Delta$ = -4.0\%, 95\% CI -5.8\% to -3.5\%, p$_{FDR}$ = 0.0469, r = 0.63, n = 7), whereas v1 ON did not yield significant changes for these muscle groups.

Similarly to abduction movement, also during flexion (Fig. \ref{fig-static2}.b) the Friedman test revealed a significant effect of assistance condition on endurance time ($\chi^2$(2) = 13.0, p=0.0015). Post-hoc Wilcoxon tests indicated that both exosuit versions significantly increased task duration relative to baseline (v1 ON: median $\Delta$ = +21.8 s, 95\% CI 7.7 s to 36.7 s, p$_{FDR}$ = 0.0410, r = 0.63, n = 8; v2 ON: median $\Delta$ = +63.5 s, 95\% CI 7.6 s to 95.5 s, p$_{FDR}$ = 0.0410, r = 0.63, n = 8), with no significant difference between the two assisted conditions (p $>$ 0.05).

Regarding muscle activation, a significant main effect of assistance condition was observed for both AD ($\chi^2$(2) = 9.8, p=0.0076) and MD ($\chi^2$(2) = 9.8, p=0.0076) activities. Post-hoc analyses revealed a muscle-specific pattern: v1 ON significantly reduced AD activation compared to baseline (median $\Delta$ = -5.5\%, 95\% CI -9.5\% to -1.9\%, p$_{FDR}$ = 0.0410, r = 0.63, n = 8), whereas v2 ON significantly reduced MD activation (median $\Delta$ = -4.3\%, 95\% CI -6.0\% to -0.8\%, p$_{FDR}$ = 0.0410, r = 0.63, n = 8). No significant differences were observed between the two exosuit versions for either muscle. For the PD, no significant main effect of assistance condition was detected ($\chi^2$(2) = 5.4, p=0.066). Nevertheless, both exosuit versions help keeping the sEMG activity low, as it already was at baseline (median MD$_{OFF}$= 4.5\%; median MD$_{v1\;ON}$= 2.5\%; median MD$_{v2\;ON}$ = 1.8\%).

Overall, both exosuit versions showed consistent trends toward reduced activation across all deltoid muscles during both movement directions. These trends did not systematically reach significance, likely reflecting the limited sample size and the relatively low baseline activations observed in this task (typically lower than 15\% MVC). 

\begin{figure}[!t]
    \centering
    \includegraphics[width=0.5\textwidth]{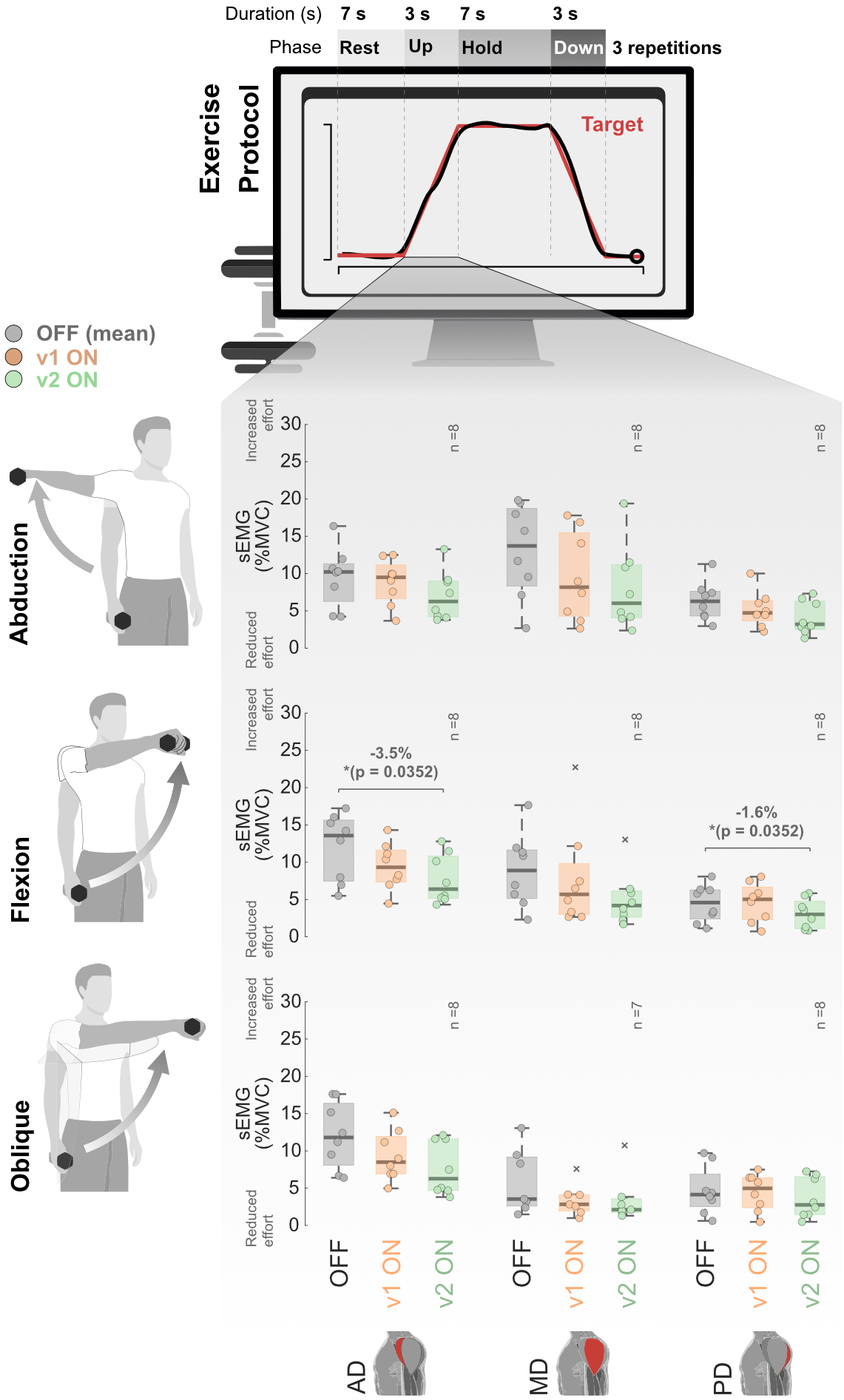}
    \caption{\textbf{Dynamic lifting task performance.} a, Normalized sEMG activity of the AD, MD and PD heads during dynamic lifting in three isolated movement directions (abduction, flexion, and oblique lift). Boxplots report group level data across unassisted (robot OFF) and assisted (v1 ON, v2 ON) conditions. Both exosuit versions show a general tendency toward reduced muscle activation during assisted movements, with version 2 yielding the most consistent reductions, particularly during forward flexion lifting.}
    \label{fig-dynamic}
\end{figure}

\subsection{Dynamic movement support is preserved in the redesigned exosuit}
Similarly to the static condition and consistently with the behavior observed for the previous design, both exosuit versions provided effective assistance during the execution of dynamic lifting movements, complementing participants’ muscular effort during the timed exercises. During shoulder abduction (Fig. \ref{fig-dynamic}), although no significant group level effects were detected, both versions showed a consistent tendency toward reduced deltoid activation, particularly for the MD, the primary muscle involved in this movement. Median MD activity decreased from baseline values of approximately 10.2\% MVC to 9.5\% MVC with v1 and 6.3\% MVC with v2. In contrast, during shoulder flexion, the Friedman test revealed a main effect for both AD (Friedman tests: $\chi^2$(2) = 10.8, p = 0.0046) and PD (Friedman tests: $\chi^2$(2) = 7.8, p = 0.0208). Post-hoc analyses showed that version 2 significantly reduced AD activation compared to baseline (median $\Delta$ = -3.5\%, 95\% CI -5.7\% to -1.9\%, p$_{FDR}$ = 0.0352, r = 0.63, n = 8), in line with its sagittal plane-oriented assistance strategy. Similarly, v2 also significantly reduced PD activity relative to baseline (median $\Delta$ = -1.6\%, 95\% CI -2.3\% to -0.4\%, p$_{FDR}$ = 0.0352, r = 0.63, n = 8). No significant differences were observed between v1 and baseline, not between the two assisted conditions. During the oblique lifting task, which required a coordinated contribution of multiple shoulder muscles, both versions were associated with a general decrease in deltoid activation, with version 2 again yielding the most pronounced reductions.

Overall, although the outcomes of the dynamic task were more distributed across muscles and movement directions than in the static condition, a consistent group level trend toward reduced muscle activation was observed with robotic assistance. The lack of systematic statistical significance across all muscles likely reflects the relatively low baseline sEMG levels, as the external load was intentionally kept moderate to avoid excessive fatigue and ensure repeatable execution.

\begin{figure*}[!t]
    \centering
    \includegraphics[width=\textwidth]{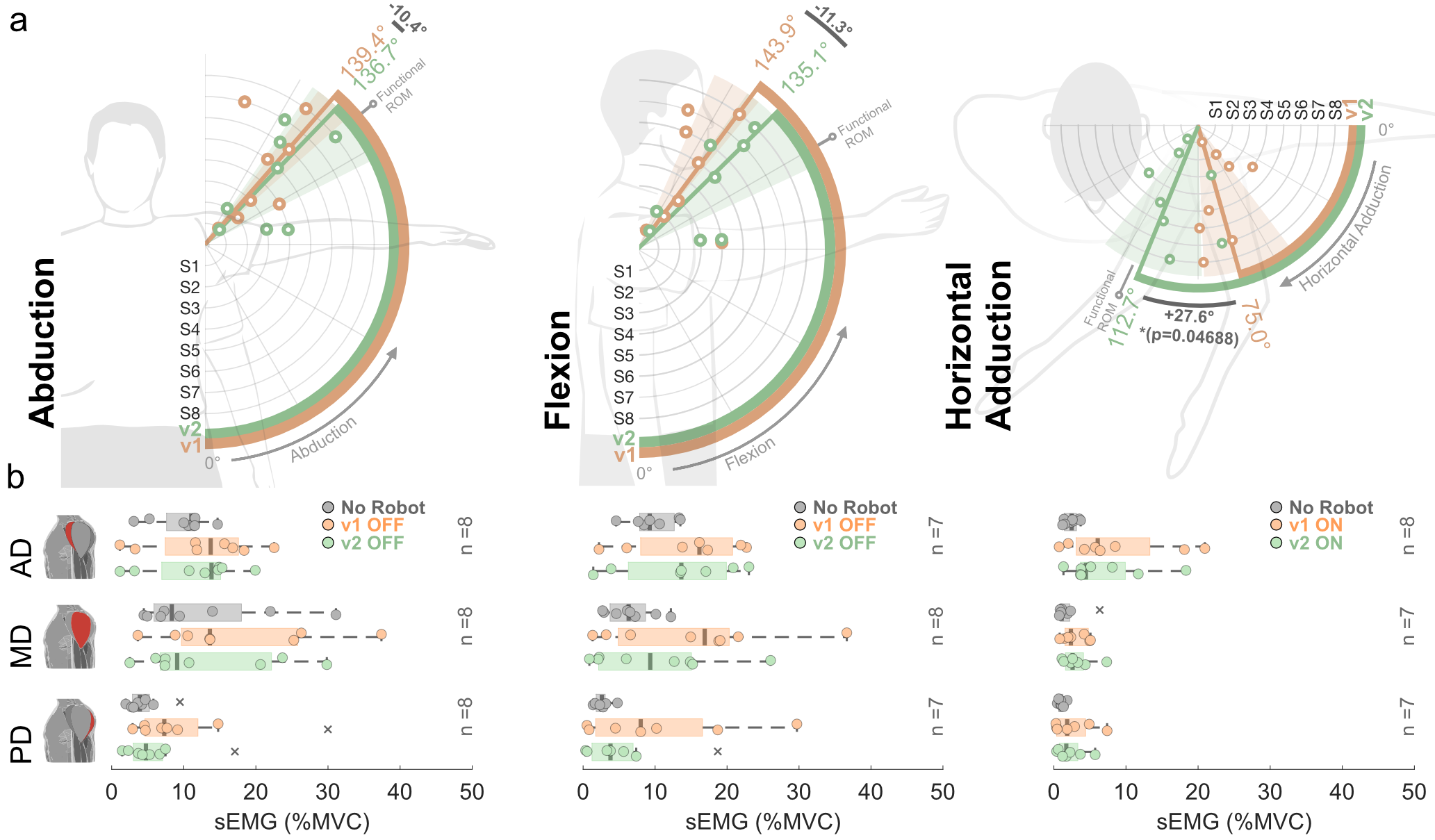}
    \caption{\textbf{Mechanical transparency during isolated shoulder movements.} a, Shoulder elevation ROM during abduction, flexion, and horizontal adduction movements. Grey lines show normative functional ROMs \cite{31} for reference. During shoulder elevation (abduction and flexion), both exosuit versions worn unpowered preserved full, physiologically relevant ROM without restricting movement. During horizontal adduction, version 2 allowed a wider shoulder azimuth ROM compared to version 1. b, Boxplots show normalized sEMG activity (\%MVC) of AD, MD, and PD heads for each movement direction. Muscle activation did not exceed the no robot baseline in any version, indicating preserved mechanical transparency.}
    \label{fig-transparency}
\end{figure*}

\subsection{Improved transparency beyond shoulder elevation}
In the transparency test we aimed to verify whether the redesigned shoulder exosuit preserves natural shoulder motion when worn unpowered and to compare its behavior with the previous versions across isolated shoulder movements (Fig. 5). For movements involving shoulder elevation (abduction, flexion), both exosuit versions demonstrated mechanical transparency. As shown in Fig. 5a, participants consistently achieved shoulder elevation angles exceeding normative functional ROM \cite{31} using either version, indicating that neither module imposed kinematic restrictions. In line with this observation, sEMG activity of the AD, MD, and PD heads remained comparable to baseline levels (Fig. 5b), with no significant increase in muscular activation when wearing either exosuit unpowered compared to the no robot baseline. These results confirm that the redesigned version 2 preserves the transparency previously demonstrated for version 1.

In contrast, differences emerged during shoulder horizontal adduction, a degree of freedom (DOF) not actively assisted by either exosuit but crucial for ADLs. A significant effect of exosuit version was observed on shoulder azimuth angle (Friedman tests: $\chi^2$(2) = 12.0, p=0.0025). While wearing version 1 significantly reduced the achievable shoulder azimuth compared to the normative ROM (median $\Delta$ = -40.0°, 95\% CI -62.0° to -27.3°, p$_{FDR}$ = 0.0469, r = 0.63, n = 8), direct comparison between the two versions confirmed a substantially larger horizontal adduction ROM with version 2 than v1 (median $\Delta$ = +27.6°, 95\% CI 14.2° to 65.9°, p$_{FDR}$ = 0.0469, r = 0.63, n = 8). Importantly, these kinematic differences were not accompanied by increased deltoid muscle activation, as no significant effects of robotic version were detected for AD, MD, or PD sEMG activity (Friedman tests, all $p >$ 0.05), indicating that the improved motion was not achieved at the expense of greater muscular effort.
Overall, these findings show that while both exosuit versions are transparent during shoulder elevation, the redesigned module markedly improves transparency along horizontal adduction by removing the kinematic constraints observed with the previous design.

\begin{figure*}[!t]
    \centering
    \includegraphics[width=\textwidth]{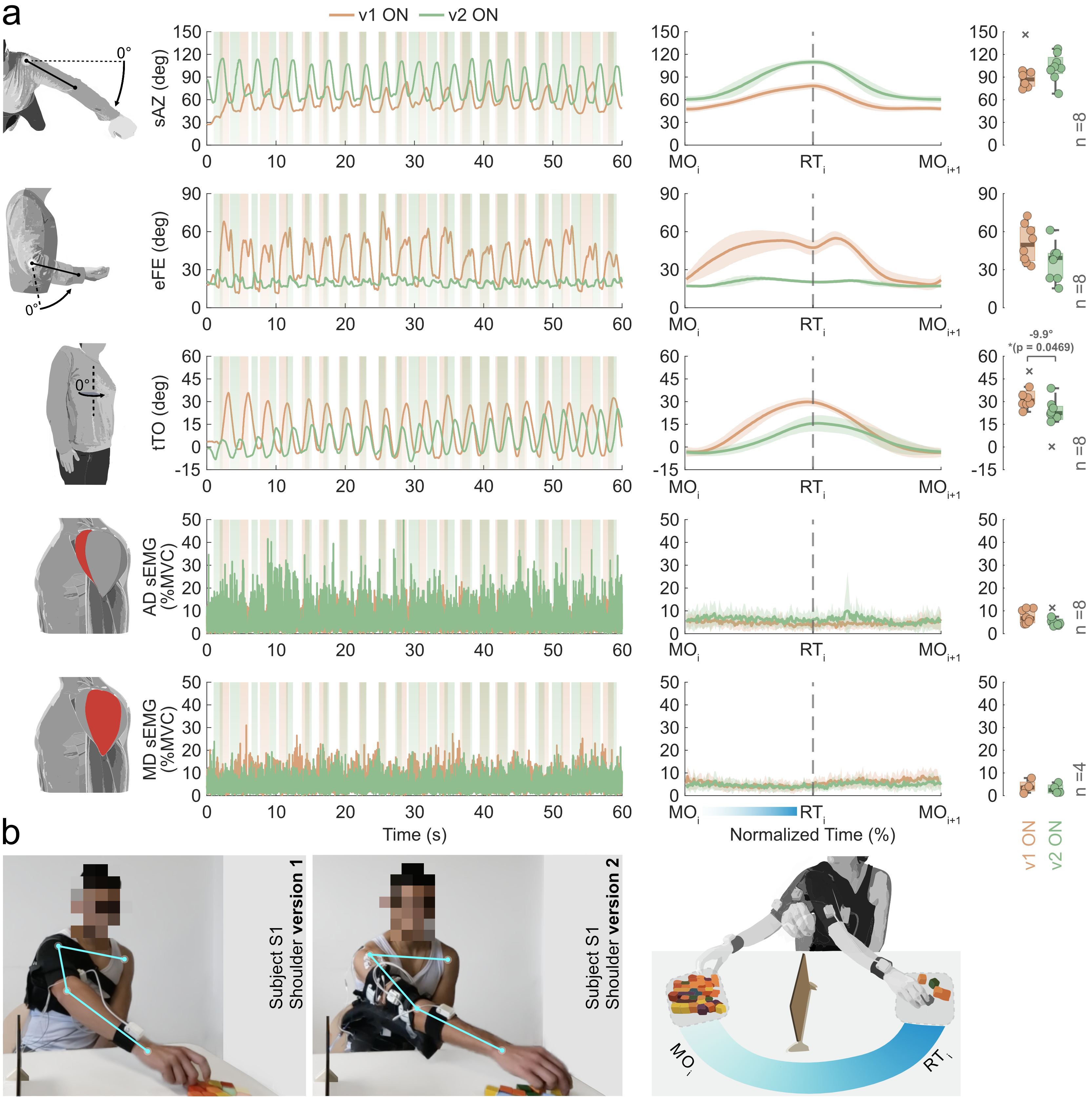}
    \caption{\textbf{Functional pick and place task highlights improved movement strategy.}  a) Kinematic and muscle activity data recorded during the pick and place task. Left: representative time-series from one participant (S1) showing shoulder azimuth (sAZ), elbow flex-extension (eFE), torso torsion (tTO), and normalized sEMG activity of the AD and MD heads during task execution with exosuit version 1 (v1 ON) and version 2 (v2 ON). Center: same signals, time-normalized and overlaid across repetitions, aligned from movement onset (MO) to release time (RT) and subsequent movement onset. Right: group level boxplots summarizing kinematic and sEMG outcomes across participants. Notably, v2 is associated with a more effective kinematic strategy, while muscle activation remains comparable across versions. b) Representative photographs of a participant performing the task with exosuit version 1 (left) and 2 (right). Compared to v1, v2 is associated with increased horizontal arm excursion, reduced elbow flexion, and decreased torso torsion during object transfer.}
    \label{fig-bbt}
\end{figure*}

\subsection{Preserved task performance with reduced compensatory movements}
During the pick and place task (Fig. \ref{fig-bbt}), both exosuit versions enabled comparable functional performance, yielding a similar number of blocks transferred within the one-minute interval (median scores: v1 ON = 19.5, v2 ON = 21.5). This indicates that the redesigned module preserved the previous module’s effectiveness during a goal-oriented activity. Although overall performance was comparable, differences emerged in movement strategies between the two versions. When using version 2, participants exhibited a larger shoulder azimuth ROM (median sAZ: v1 ON = 87.0°, v2 ON = 102.8°) and a reduced elbow flexion angle compared to version 1 (median eFE: v1 ON = 49.6°, v2 ON = 39.3°), suggesting increased horizontal shoulder involvement and reduced reliance on more distal joints during object transfer. Importantly, version 2 was associated with a significantly lower torso torsion angle than version 1 (median $\Delta$ = -9.9°, 95\% CI -16.8° to -3.6°, p$_{FDR}$ = 0.0469, r = 0.63, n = 8), indicating reduced engagement of compensatory trunk movements. Deltoid muscle activation remained comparable between the two versions, confirming that the observed kinematic differences were not achieved at the expense of increased muscular effort.

Overall, these findings indicate that while both exosuit versions effectively supported functional manipulation, the redesigned module promoted a more efficient and natural movement strategy, characterized by enhanced shoulder mobility and reduced compensatory involvement.

\begin{figure*}[!t]
    \centering
    \includegraphics[width=\textwidth]{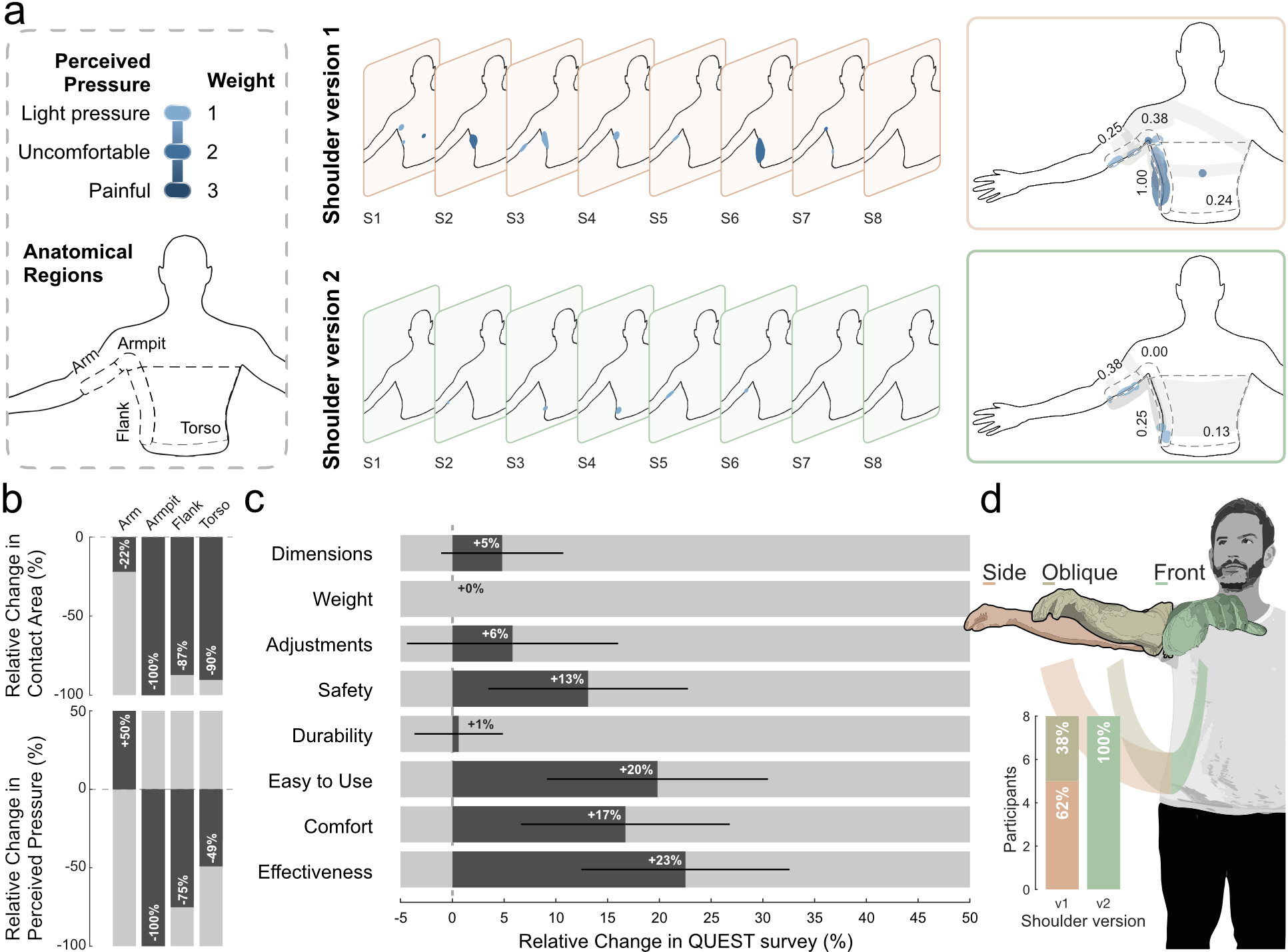}
    \caption{\textbf{User-reported comfort and perceived interaction.} a)  Left: legend relating reported sensation levels to intensity weights, and definition of anatomical regions on the torso template. Center: individual maps of perceived contact pressure for each exosuit version. Right: group level average weighted intensity maps, highlighting differences in contact area and pressure distribution. b)  Regional relative changes in contact area and perceived pressure intensity across anatomical regions, indicating an overall reduction in high-intensity pressure with version 2. c)  Relative variation in post-experiment QUEST scores (mean ± SEM) between exosuit versions, showing markedly improved ratings for version 2 in comfort, ease of use and perceived effectiveness. d)  Perceived primary direction of assistance reported by participants, revealing a shift toward frontal assistance with version 2, consistent with its redesigned anchoring and support strategy.}
    \label{fig-comfort}
\end{figure*}

\subsection{Redesigned exosuit improves user comfort and perceived interaction}
User-reported comfort evaluation revealed clear differences between the two exosuit versions. Group-level weighted pressure maps (Fig. \ref{fig-comfort}.a, Fig. \ref{fig-comfort}.b) showed a reduced overall contact area with version 2, together with a shift toward lower perceived pressure intensity, particularly in sensitive regions such as the armpit and flank. An exception was observed at the upper arm region, which exhibited the smallest reduction in contact surface and a slight increase in weighted pressure score; however, this change remained modest (from 0.25 to 0.38) and well below the level associated with uncomfortable sensations.
Post-experiment user satisfaction scores (QUEST) consistently improved from version 1 to version 2 (Fig. \ref{fig-comfort}.c). Participants reported higher satisfaction for ease of use (+20\%), comfort (+17\%), and perceived effectiveness (+23\%), alongside positive changes in safety (+13\%) and other product-related items (e.g., adjustability and dimensions). Ratings related to weight and durability remained largely unchanged: the former likely reflecting an already well-accepted lightweight design typical of soft robotic systems, and the latter representing a qualitative impression that would require longer-term and real-world use to be more thoroughly assessed.

Regarding perceived direction of assistance, participants predominantly identified version 2 as providing frontal assistance, whereas version 1 was more frequently associated with lateral or oblique support. This perceptual shift is consistent with the redesigned anchoring and sagittal plane-oriented assistance strategy of the updated module. 

\section{DISCUSSION}
In this study, we evaluated whether a redesigned shoulder exosuit could enhance the assistive capabilities of our previous module, especially in terms of comfort and functional interaction with the user. Across all tasks, the two versions showed a strong overlap in core performance: both effectively supported physically demanding exercises. In particular, during static holding exercises both designs increased endurance time, while reducing deltoid activation. During dynamic lifting, robotic support complemented muscular effort across movements, with effects aligned with each version’s target plane of motion. Transparency tests further confirmed that version 2, similarly to version 1, did not constrain shoulder elevation when unpowered. 

Beyond preserving performance, the redesigned module demonstrated clear functional advantages. By shifting assistance toward the sagittal plane, version 2 reduced the effort required to position the arm forward without increasing muscular demand, which could translate into more comfortable, effective and natural support to arm movements involved in common ADLs (e.g., manipulation, feeding, self-care). Importantly, while the redesigned module primarily assists shoulder flexion, it preserves transparency in horizontal abduction-adduction, allowing users to abduct the arm with a natural level of effort. This aspect is particularly relevant given that everyday functional tasks rely on coordinated multi-joint movements rather than isolated DOFs, and often require substantial elbow flexion to effectively position the hand within the workspace \cite{1}. Moreover, version 2 also preserved horizontal plane mobility, allowing larger shoulder azimuth ranges without increasing muscle activation. This confirms the improved kinematic behavior was not achieved at the expense of higher muscular effort.

Differences between the two versions became most evident during functional tasks. In the pick and place activity, both exosuits enabled comparable task performance, but version 2 promoted a distinct movement strategy characterized by increased shoulder joint excursion, reduced elbow flexion, and significantly lower torso torsion. The latter indicates reduced reliance on compensatory motor behavior, which is commonly observed in stroke survivors with upper-limb impairments \cite{32}. Although tested in unimpaired participants, these results suggest that the redesigned module facilitates a more natural and efficient coordination of upper-limb movements, thus again a more comfortable support.

Consistent with the primary objective of this study, user-reported outcomes highlighted a substantial enhancement in wearability with the redesigned module. Weighted pressure maps revealed an overall reduction in contact area and a shift toward lower perceived pressure intensity. Notably, despite a purposeful increase and local stiffening of the contact interface, particularly on the flank through the thermoplastic rigid plate, participants reported lower perceived pressure in this region. This likely reflects a more uniform distribution of contact forces over a larger effective area, which has been previously associated with improved perceived comfort in wearable interfaces \cite{21}. Although a modest increase in perceived pressure was observed at the arm, this remained well below discomfort levels. Notably, four out of eight participants reported uncomfortable sensations when using version 1, whereas no participant reported discomfort with version 2, underscoring the importance of participant-level acceptability for wearable assistive devices, particularly in scenarios involving prolonged or daily use. Post-experiment satisfaction scores further reflected improvements in ease of use, comfort, and perceived effectiveness for version 2, while attributes such as weight and durability remained unchanged, likely reflecting an already favorable baseline perception of soft robotic systems. Importantly, these results contribute to bridging the gap between the development of effective exosuits and the need to more explicitly account for user-centered aspects of human-robot interaction, in line with recent calls within the field \cite{12}.

This study was designed as a qualitative, comparative evaluation rather than a large-scale quantitative validation. Given the sample size, conservative non-parametric statistics were adopted, and the focus was placed on relative differences between versions rather than absolute performance. No dedicated mechanical characterization was performed; however, the underlying actuation technology closely matches that of the previous design, for which these analyses already exist \cite{24,25}.

Assessment of muscle activity was limited to the deltoid muscles, excluding additional muscles that could further inform transverse plane behavior, a deliberate choice to prioritize participant comfort. Finally, although the comfort-mapping framework relied on simplified assumptions – namely, a linear weighting of perceived pressure levels and uniform cutaneous sensitivity across anatomical regions – which were adequate for the present comparative analysis, it proved valuable as a tool to capture perceived, rather than purely measured, human-robot interaction. Beyond this study, such an approach could be further formalized, and proposed as a complementary method to support the systematic characterization of user experience in wearable assistive technologies and to guide future design iterations.
Overall, these results show that the targeted, user feedback-driven design refinements can yield meaningful improvements in functional behavior and comfort without compromising assistive performance. Future work will focus on validating intent-driven control strategies, extending testing to clinical populations, and advancing toward an effective, comfortable, fully portable multi-module wearable system.

\section*{APPENDIX}

\begin{table}[!h]
    \centering
    \caption{Demographics data of the participants.}
    \begin{tabular}{cccccc}
    ID   & Weight   & Height   & Age     & Sex     & Handedness \\
         & kg & cm & y & - & - \\
        \midrule
    S1   & 73.5     & 176      & 26      & M       & right      \\
    S2   & 52.0     & 162      & 21      & F       & right      \\
    S3   & 83.0     & 179      & 25      & M       & left       \\
    S4   & 72.0     & 185      & 28      & M       & right      \\
    S5   & 75.0     & 182      & 25      & M       & right      \\
    S6   & 58.0     & 157      & 23      & F       & right      \\
    S7   & 54.0     & 159      & 27      & F       & right      \\
    S8   & 50.0     & 165      & 31      & F       & right      \\
         \midrule
    Mean & 64.7     & 170.6    & 25.8    &     -    &      -      \\
    Std  & 12.6     & 11.1     & 3.1     &      -   &       -    
    \end{tabular}
    
\end{table}


\end{document}